# Large Language Models or AI Agents for English to Arabic Translation? An assessment on WikiEn-Ar190

**Fatima Zahra Iguenfer**
Al Akhawayn University in Ifrane
f.iguenfer@aui.ma

**Yousra Chtouki**
Al Akhawayn University in Ifrane
y.chtouki@aui.ma

## Abstract

This research investigates whether a large language model or an autonomous AI Agent performs better at translating from English to Arabic, which is considered a low-resource language. The authors introduce WikiEn-Ar190, a new dataset of 190 carefully selected Wikipedia articles from 18 different areas, which were created and verified by two native Arabic speakers to ensure that they accurately reflect both meaning and style. Using this dataset, the study evaluates the translation quality of GPT-5-mini both as an independent neural translator and when used within a more complex process that includes self-assessment and improvement. As a standalone model, GPT-5-mini produced very fluent translations , achieving a COMET score of 0.8649 and receiving high human scores for adequacy, fluency, and style (averaging 4.53 out of 5). When used in a system that combined translation with self-assessment, it improved lexical accuracy by 28.76% (BLEU score of 22.83) while maintaining a similar level of semantic quality. These findings show that methods involving autonomous reasoning can improve translation consistency for language pairs with limited resources. Beyond the results, the approach offers a framework for evaluating and improving translations, potentially using them as assisting tools to expand the representation of low resource languages, such as Arabic, on the web.

## 1 Introduction

Machine translation (MT) has experienced considerable progress with the emergence of large language models (LLM), particularly with models such as GPT-4 and Google Translate establishing themselves as frontrunners, outperforming traditional systems in terms of fluency and cultural adaptation and achieving human-level translations (Yan et al., 2024a). The recent debut of GPT-5 in August 2025 marks the introduction of a new model generation, although its capabilities for English-to-Arabic (EN-AR) translation remain largely unexplored. Investigating such advanced systems requires robust parallel corpora, which are limited for low-resource languages like Arabic, thus constraining the development and evaluation of state-of-the-art MT technologies. Wikipedia itself highlights the difference in the number of articles available in Arabic compared to English, as evidenced by the count of articles in both languages shown in Figure 1.

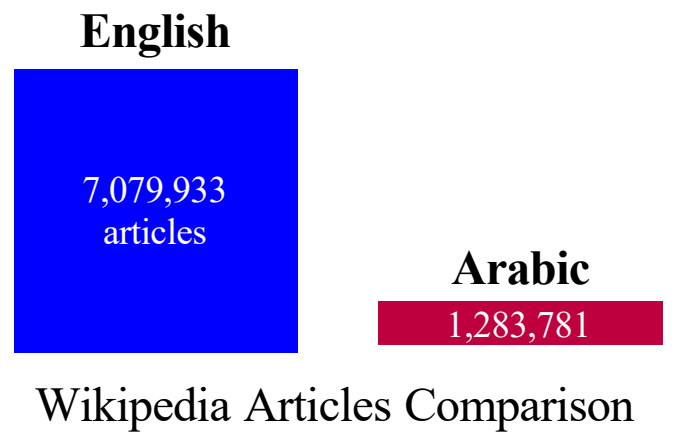

Figure 1: Comparison of article counts in English and Arabic Wikipedia (as of October 25, 2025).

Nevertheless, Wikipedia remains a valuable, yet underutilized, resource despite challenges such as non-native translations. This study introduces WikiEn-Ar190, an innovative parallel corpus comprising 190 Wikipedia excerpts, and, to our knowledge, is the first to evaluate GPT-5-mini as both an independent LLM and within an AI Agent framework. The high quality of the generated translations indicates its potential for translating English content into Arabic or serving as a computer-assisted translation (CAT) tool.

The remainder of this paper is structured as follows: Section 2 surveys the related work on EN-AR MT, LLM, and agentic AI, providing a backdrop to current challenges and methods. Sections 4 and 3 detail the dataset and evaluation systems, including GPT-5-mini and the AI Agent, describing the data collection, prompt design, and evaluation methodologies. Finally, Section 5 concludes the paper with a summary of findings and directions for future research.

## 2 Related Work

The Arabic language poses different linguistic challenges due to its complex morphology. English ↔ Arabic machine translation (MT) has progressed from rule-based and example-based MT to statistical and neural approaches and then to the exploration of large language models (LLM). Despite advances, English ↔ Arabic MT remains underdeveloped due to limited parallel corpora, with LLMs showing promise but lagging in low-resource languages compared to high-resource ones. Recent research also investigates prompt engineering to enhance LLM translation quality and introduces agentic AI systems, which use autonomous multi-agent workflows to potentially outperform traditional NMT, though their effectiveness has not been extended to low-resource languages.

### 2.1 Context on English ↔ Arabic Machine Translation

Machine translation (MT) between English and Arabic (English ↔ Arabic) is particularly challenging due to the significant linguistic differences between the two languages. Arabic is characterized by its rich morphology, involving complex derivational processes, non-concatenative morphology, and a root-and-pattern system that is fundamentally different from that of English (Habash, 2010; Soudi et al., 2007). The evolution of machine translation technologies used for English ↔ Arabic translation has progressed through several paradigms, each of which tries to address these challenges with an increasing degree of sophistication. First, early rule-based machine translation systems (RBMT) emerged and relied on linguistic knowledge encoded by expert linguists (Hutchins, 1995). Then a shift was seen towards Example-based Machine Translation (EBMT), where a large collection of pre-translated example sentences is used to translate new Arabic text. Given the scarcity of Arabic corpora, this MT technique has failed in several research works (Ehab et al., 2018). To overcome the limitations of both RBMT and EBMT, researchers adopted hybrid approaches to tackle challenges such as agreement and ordering which come from the fact that different languages have different text orientations and order of words (Alawneh and Sembok, 2011). Following all these paradigms, Statistical Machine Translation (SMT) gained traction. Instead of relying on linguistic knowledge, SMT, a supervised machine learning problem, uses statistical models to convert text from one language to another. However, it still relies on extensive parallel text corpora that are rarely found in low-resource languages such as Arabic. Finally, Neural Machine Translation (NMT) especially Large Language Models (LLMs) have changed how machine translation is perceived. However, European languages have received the most attention from this revolution, with English ↔ Arabic MT systems still stuck in their early stages due to the shortage of parallel corpora.

### 2.2 LLMs for English ↔ Arabic Machine Translation

LLMs such as GPT have been proven to produce a highly fluent and competitive translation output even in a zero shot setting (Hendy et al., 2023). It highly competes with commercial translation tools like Google Translate for high-resource European languages, but it significantly falls behind when it comes to low-resource or less common languages (Jiao et al., 2023). Research has found that NMT systems like GPT and Google Translate lag in translating from English to Arabic. They require extensive study and detailed analysis to understand the types of errors that occur in various text forms (Alzain et al., 2024). Although English ↔ Arabic MT is still immature, research on other pairs of high-recourse languages is investigating how to improve the quality of LLM translation through prompt engineering. When the information is not too complex or noisy, ChatGPT performs better than other commercial systems with additional information, such as translation tasks and context domain (Gao et al., 2023a). In a specific study on English to Arabic translation, researchers found that using translated sentences with the highest error on prompting improved the performance of the GPT model for tanslation tasks (Nagi et al., 2024). However, further investigations on the efficiency of informed prompts should be conducted beyond high-recourse languages.

### 2.3 Agentic AI for English ↔ Arabic Machine Translation

An artificial intelligence agent (AI Agent) can be defined as a system capable of executing tasks autonomously by conceptualizing workflows using accessible tools. These agents exhibit reasoning, planning, and memory capabilities, possessing a degree of autonomy that allows them to make decisions, learn, and adapt to new circumstances. One of the first works on the use of AI Agent systems in MT is the research conducted by Vicent Briva-Iglesias (Briva-Iglesias, 2025). The findings show that multi-agent workflows obtain higher translation quality than traditional NMT systems and/or single-agent systems in MT tasks. On the other hand, a different study by the same author indicates that Agentic workflows exhibit considerable variation between language pairs, implying that their success is largely reliant on factors specific to the task. (Iglesias and Dogru, 2025). Given the novelty of the field, it has not been tackled by Arabic ↔ other language pairs; a significant research gap in low-resource language applications.

In this work, not only do we contribute to expanding Arabic language resources through the WikiEn-Ar190 dataset, but we also position our paper as the first to assess the potential of an LLM such as GPT-5-mini to provide accurate and fluent Arabic translations from English texts. Both when used as a standalone NMT system and when integrated into an AI Agent workflow.

| **Domain** | **Count** | **%** |
|---|---|---|
| History | 65 | 34.2 |
| Culture & Society | 26 | 13.7 |
| Novels | 19 | 10.0 |
| Engineering & Technology | 15 | 7.9 |
| Culture & Society Bio. | 14 | 7.4 |
| Autobiographies | 8 | 4.2 |
| Health & Medicine | 6 | 3.2 |
| Literature & Theatre | 6 | 3.2 |
| Education | 5 | 2.6 |
| Computing | 4 | 2.1 |
| Magazines & Comics | 4 | 2.1 |
| Media | 4 | 2.1 |
| Law Biographies | 4 | 2.1 |
| Business & Economics Bio. | 3 | 1.6 |
| Law | 3 | 1.6 |
| Heraldry & Honors | 2 | 1.1 |
| Mathematics | 1 | 0.5 |
| Mathematics Bio. | 1 | 0.5 |
| **Total** | **190** | **100.0** |

Table 1: Dataset distribution across fields

## 3 Dataset

### 3.1 Data Collection

We introduce WikiEn-Ar190, a parallel English-Arabic dataset derived from Wikipedia articles. We selected Wikipedia articles because they are commonly employed as pre-training datasets for numerous Natural Language Processing (NLP) endeavors, particularly for low-resource languages such as Arabic. This is attributed to Wikipedia's extensive multilingual content and a rich set of metadata that can be measured and contrasted (Mittermeier et al., 2021). However, it is important to note that not all Arabic Wikipedia entries are written by native Arabic speakers. In reality, a significant number of these articles have been automatically generated using commercial tools such as Google Translate. For this reason, WikiEn-Ar190 was constructed through a systematic multistage process. We manually identified Wikipedia articles with parallel Arabic versions and selected passages where the Arabic translation demonstrated high accuracy. Two native Arabic-speaking annotators with expertise in computational linguistics reviewed each selected passage. The annotators corrected translation errors, checked for semantic equivalence, and replaced English words with appropriate Arabic equivalents, except for numerical expressions and dates which remained in Latin script.

### 3.2 Dataset Characteristics

The dataset spans 18 fields to ensure a broad domain coverage, as shown in Table 1. It also includes text segments with an average length of 116.5 words for English ($\sigma = 47.4$) and 105.0 words for Arabic ($\sigma = 45.0$). English texts range from 30 to 259 words, while the Arabic texts range from 30 to 264 words. The complete corpus contains 22,126 English words and 19,955 Arabic words. This length distribution ensures that each segment provides sufficient context for a meaningful translation evaluation while remaining manageable for detailed analysis.

## 4 Systems Under Evaluation

### 4.1 Large Language Model (GPT-5-mini)

In this experiment, we employ **GPT-5-mini** as our LLM translator. The release of GPT-5 (August 2025) marks a new generation of models that, to the best of our knowledge, has not yet been evaluated for English-to-Arabic machine translation (MT). Previous research has shown that GPT-4 performs comparably to junior human translators, although its translation capability gradually weakens when moving from resource-rich to resource-poor language pairs (Yan et al., 2024b). GPT-5 has been introduced as a more robust model than its predecessor GPT-4, thus we expect higher quality translations. We selected GPT-5-mini from the GPT-5 series because of its cost-effectiveness, which balances performance with resource efficiency.

**Building the prompt.** Recent research has shown that ChatGPT can achieve superior results compared to commercial translation systems such as Google Translate when properly prompted (Gao et al., 2023b). According to the ISO 17100 translation standard, high-quality professional translation depends on clearly specifying the purpose, target audience, and style before the translation process (ISO, 2015). In line with this, Yamada (Yamada, 2023) examined the effects of incorporating these preproduction parameters into prompts for Japanese ↔ English translation using ChatGPT. The study found that explicitly defining the purpose and audience of the translation significantly improved the quality of the output, resulting in translations that were more culturally appropriate, stylistically natural, and aligned with professional standards compared to conventional MT systems.

Furthermore, the format and wording of a prompt directly affect how LLMs interpret and execute translation tasks (Gao et al., 2023c). To design our translation prompt, we follow the findings of (Gao et al., 2023c), which demonstrate that ChatGPT's translation quality can be substantially enhanced by explicitly providing contextual information. Specifically, the authors highlight the importance of including:

- *Translation task information* (e.g., English-to-Arabic)
- *Context domain information* (e.g., Wikipedia articles and associated fields)
- *A few high-quality example translations* (1–5 shots).

Based on these insights, we developed the prompt structure shown in Figure 2, which integrates the purpose of translation, the audience, the language pair, and the domain context. Finally, we include five example translation pairs (shots) drawn from outside the dataset, but still corresponding to Wikipedia articles, to guide the model toward a better output.

*Extracted from the field column of the dataset.

**Preliminary Evaluation.** To evaluate the performance of GPT-5 mini, we split WikiEn-Ar190 into 80% training set (152 texts) and 20% testing set (38 texts). This split ensures a holdout test set for an unbiased final evaluation, while the training set was used for preliminary analysis, following standard MT practices to avoid data leakage. After translating the training set using the prompt in Figure 2, a manual review by our annotators indicated excellent quality in terms of accuracy and fluency, although translations often diverged from the reference texts at the word level. To quantify this, we computed the BLEU and COMET scores in a subset of 25% of the training data (38 texts); the results are shown in Table 2

> You are an expert translator specializing in English to Modern Standard Arabic translations. Your goal is to produce high-quality, accurate translations that are culturally appropriate, stylistically natural, and aligned with professional standards as per ISO 17100.
>
> **Purpose:** Produce translations of encyclopedic Wikipedia excerpts for evaluation in a machine translation research.
> **Target Audience:** Arabic-speaking general readers.
> **Style:** Encyclopedic.
> **Language Pair:** English to Modern Standard Arabic.
> **Domain Context:** The text is from a Wikipedia article in the field of {field}*.
> Here are five high-quality example translations to guide your output: [Example 1] ... [Example 5]
>
> Output only the Arabic translation without any additional explanations or text.

Figure 2: Prompt for English ↔ Arabic Translation with Contextual Guidance.

| Metric | Score |
|---|---|
| BLEU | 18.84 |
| COMET | 0.8579 |

Table 2: Preliminary Evaluation Metrics on a 38-Text Subset of the Training Data

The low BLEU score, contrasted with the high COMET score and manual assessment, supports our hypothesis that BLEU penalizes valid lexical variations in Arabic due to its focus on n-gram overlap.

**Human Evaluation** To validate the translation quality of GPT-5 mini beyond the metrics, we performed a human evaluation with the help of our two native Arabic-speaking annotators in the same 38-text subset of the training data previously assessed with BLEU and COMET.

Following established English ↔ Arabic MT evaluation protocols, both annotators indepen-dently rated each translation on three dimensions using a 1-5 Likert scale, aligned with ISO 17100 professional translation standards.

| Annotator | Adequacy | Fluency | Style |
|---|---|---|---|
| Annotator 1 | 4.55 | 4.53 | 4.47 |
| Annotator 2 | 4.58 | 4.50 | 4.53 |
| Average | 4.57 | 4.52 | 4.50 |

Table 3: Human Annotators Evaluation Scores on a 38-Text Subset of the Training Data with Averages

| Metric | Cohen's Kappa |
|---|---|
| Adequacy | 0.82 |
| Fluency | 0.80 |
| Style | 0.76 |

Table 4: Inter-Annotator Agreement (Cohen's Kappa) for Human Evaluation on a 38-Text Subset of the Training Data

- **Adequacy:** How well does the translation convey the meaning of the English source text?
- **Fluency:** How natural and grammatically correct is the Arabic translation for a native Arabic speaker?
- **Style:** How well does the translation maintain the encyclopedic/Wikipedia style and tone of the source?

Scores were collected using a structured Excel template containing source texts, GPT-5 mini translations, and reference translations, allowing per-text assessment followed by corpus-level aggregation. The results are shown in Table 3

The average corpus-level score of 4.53 (out of 5) in Adequacy, Fluency, and Style confirms high translation quality and supports our hypothesis that low BLEU scores reflect lexical variations in Arabic. To further assess the consistency of human evaluations, we calculate Cohen's Kappa to measure the agreement between our two annotators in the 38-text subset for each metric (Adequacy, Fluency, Style). The Kappa results are reported in Table 4.

With the high-quality translations validated in the training subset, we proceeded to evaluate the performance of GPT-5 mini in the held test set. The results of the test set, shown in Table 5, demonstrate consistent performance, with a BLEU score of 17.73 and a COMET score of 0.8649, further supporting the effectiveness of GPT-5 mini despite lexical divergence from reference texts.

| Metric | Score |
|---|---|
| BLEU | 17.73 |
| COMET | 0.8649 |

Table 5: Evaluation Metrics for GPT-5 mini on the Test Set

#### 4.1.1 AI Agent

To combine both the translation and the evaluation processes, we developed an AI agent that integrates the roles of translator and evaluator, previously performed by human annotators and the GPT-5-mini model, into a single system. Using GPT-5-mini, the agent first translates the English Wikipedia excerpts into Modern Standard Arabic with the enhanced prompt shown in Figure 2. It then acts as a critical Arabic linguist, evaluating its own translations based on the scoring criteria (Adequacy, Fluency, Style) defined in Section 4.1. Finally, the agent refines the translations based on its self-assessment when improvements are needed. The AI Agent workflow, shown in Figure 3, uses a chain-of-thought approach to enhance translation quality through iterative reasoning (Wei et al., 2023).

**Prompt Design.**

*Prompt 1: Initial Translation.* We reuse the translation prompt from Figure 2 to generate the initial English to Arabic translation.

*Prompt 2: Self-Evaluation.* To ensure a self-contained agent, we exclude reference texts. The prompt, shown in Figure 4, evaluates translations based on the criteria in Section 4.1.

*Prompt 3: Refined Translation.* The prompt in Figure 5 refines the translation based on the self-evaluation critique, to ensure high adequacy, fluency, and encyclopedic style.

The AI Agent uses a chain-of-thought approach, integrating three prompts to improve translation quality through iterative reasoning (Wei et al., 2023). The process begins with an initial translation from English to Arabic, followed by an initial self-evaluation of adequacy, fluency, and style, the results aggregated in Table 6.

Using the self-evaluation critique, the agent refines translations to improve quality. The final translations are evaluated using the BLEU and

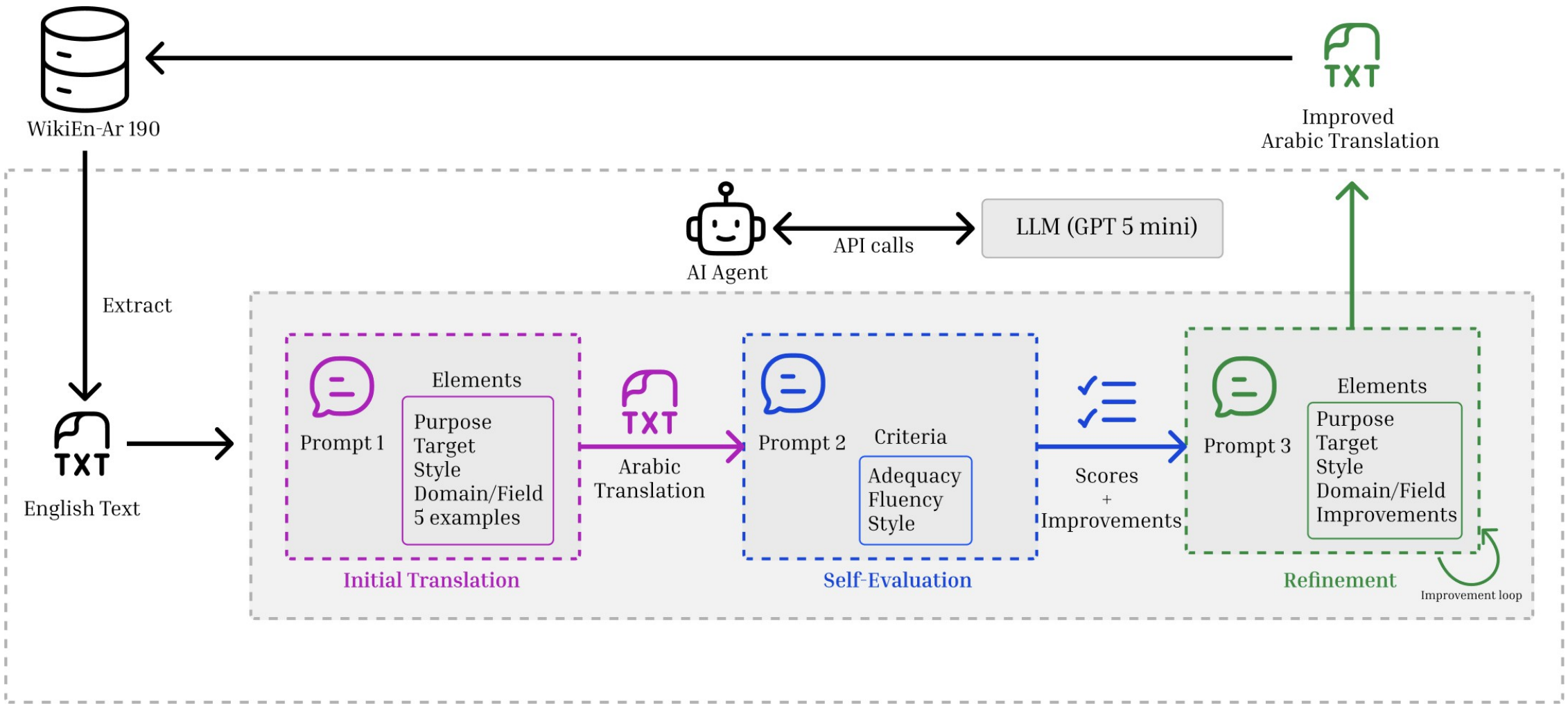

Figure 3: AI Agent workflow

You are an expert translator specializing in English to Modern Standard Arabic translations. Your task is to evaluate LLM-generated translations of Wikipedia excerpts for machine translation research. Assess the translations on:
Adequacy: Does the translation convey the source text's meaning ? (5=perfect, 1=unacceptable)
Fluency: Is the Arabic translation natural and grammatically correct ? (5=perfect, 1=unacceptable)
Style: Does the translation maintain the encyclopedic style and tone? (5=perfect, 1=unacceptable)
Provide scores (1–5) for each metric.

**Input:** English Text: {english_text} Translation: {translation} Field: {field}
**Output format:**
Adequacy: [score]/5
Fluency: [score]/5
Style: [score]/5
Improvements: [Improvement 1] [Improvement 2, if applicable] [Improvement 3, if applicable]

Output only the scores and improvements without additional text.

Figure 4: Prompt for self-evaluation of Arabic translations.

| Model | Adequacy | Fluency | Style |
|---|---|---|---|
| GPT-5 Mini | 4.82 | 4.82 | 4.82 |

Table 6: Average self-assessed scores for the AI Agent (GPT-5 Mini) on the test set.

COMET metrics, reported in Table 7.

To evaluate the effectiveness of our designed AI Agent workflow, we compare its performance to GPT-5 Mini standalone translations, as shown in Table 8. The BLEU score improved from 17.73 to 22.83 (+28.76%), while the COMET score decreased slightly from 0.8649 to 0.8629 (-0.23%).

You are an expert translator specializing in English to Modern Standard Arabic translations. Refine the provided Arabic translation based on the critique, confirming high adequacy, fluency, and encyclopedic style for Arabic Wikipedia. Focus solely on the suggested improvements.

**Input:** Original Translation: {translation} Critique: {critique} Field: {field}
**Output:** Refined Arabic translation only.

Figure 5: Prompt for Arabic translations Refinement

| Metric | Score |
|---|---|
| BLEU | 22.83 |
| COMET | 0.8629 |

Table 7: Evaluation metrics for the AI Agent's refined translations on the test set.

| Model | BLEU | Impr. (%) | COMET | Impr. (%) |
|---|---|---|---|---|
| GPT-5 Mini | 17.73 | – | 0.8649 | 0.23 |
| AI Agent | 22.83 | 28.76 | 0.8629 | – |

Table 8: Comparison of BLEU and COMET scores for LLM (GPT-5 Mini) and AI Agent

Our explanation suggests that the results are due to the agent translation process. In fact, the increase in BLEU indicates better lexical overlap and n-gram overlap, suggesting improved word choice and syntax for Arabic, influenced by the agent's self-evaluation of adequacy, fluency, and style. However, the minor drop in COMET reflects possible semantic shifts, as COMET measures meaning preservation. Without reference texts, the agent focused on style and fluency, improving BLEU but sometimes diverging from semantic nuances, affecting COMET.

## 5 Conclusion

This research explored whether advanced language models like GPT-5-mini or agentic AI systems deliver better machine translation quality from English to Arabic. We assessed both methods using WikiEn-Ar190, a parallel corpus we developed consisting of 190 Wikipedia excerpts covering 18 different fields. Our results indicate that while GPT-5-mini alone delivers translations with high semantic accuracy (COMET score of 0.8649) and strong fluency as verified by human assessors (average scores of 4.53 out of 5), its word choices often deviate from the reference texts, leading to moderate BLEU scores (17.73). The agentic AI process, which includes translation, self-assessment, and iterative refinement through chain-of-thought reasoning, led to a 28.76% increase in the BLEU score while maintaining similar semantic quality (COMET 0.8629). This enhancement implies that autonomous multi-agent systems can enhance translation quality for low-resource language pairs by systematically addressing adequacy, fluency, and stylistic uniformity.

The importance of this study extends beyond improvements in metrics. It demonstrates that agentic AI workflows can be effectively utilized for low-resource languages like Arabic, filling a gap in machine translation research that has largely concentrated on high-resource European languages. Moreover, WikiEn-Ar190 offers the research community a high-quality benchmark dataset for evaluating English-to-Arabic translation systems.

## 6 Limitations

This study has a number of limitations that need to be recognized. Firstly, we did not incorporate an iterative improvement loop at the refiner stage, where translations are refined until reaching a specific average score threshold, which might result in higher quality outputs. Secondly, we did not conduct comparisons of our results with well-established commercial systems such as Google Translate, which are known for their effective performance in English-to-Arabic translation. Nonetheless, our main objective was to showcase the potential of LLMs within agentic workflows as computer-assisted tools for expanding existing English texts and translating them into Arabic, rather than offering comprehensive benchmarking.